\documentclass[5p]{elsarticle}
\usepackage{hyperref}
\usepackage{xurl}
\usepackage{enumitem}
\usepackage{adjustbox}
\usepackage{subcaption}
\usepackage[skip=2pt]{caption} % Adjust the 'skip' value as needed
\usepackage{natbib}
\begin{document}
\begin{frontmatter}

\title{Analysing drivers and interdependencies in European electricity markets using XAI}
\author{Antoine Pesenti\corref{cor1}}
\author{Aidan O'Sullivan}
\cortext[cor1]{Corresponding author: antoine.pesenti.21@ucl.ac.uk}
\address{UCL Energy Institute, University College London, UK}
\date{June 2026}

\begin{abstract}
Electricity markets are inherently complex systems characterised by strong nonlinearities, high-dimensional interactions, and increasing interdependence across regions. While deep neural networks (DNNs) have demonstrated strong predictive capabilities for electricity prices, their lack of interpretability limits their usefulness for understanding the underlying drivers of price formation.\par
This paper addresses this gap by combining DNN models with explainable artificial intelligence (XAI) techniques to analyse the determinants of electricity prices across 39 European bidding zones. We employ SHAP (SHapley Additive exPlanations) to quantify feature contributions and apply and extend SSHAP, an aggregation framework to improve interpretability in high-dimensional settings.\par
The analysis identifies that renewable energy sources, particularly solar, play a disproportionately important role in price formation despite their lower share in total power generation. Gas prices remain a dominant and consistent driver across electricity markets, while interconnections significantly shape price dynamics, highlighting the strong interdependence of European electricity systems. In addition, a synthetic EU-wide electricity market is constructed to explore the counterfactual scenario of a fully integrated market with a single price.\par
\end{abstract}

\begin{highlights}
\item 39 European electricity markets are analysed with XAI methods applied to their DNN models
\item An EU-wide electricity market is set to explore the scenario "What if there was a common price in the European market"
\item While solar power contributes the least to overall electricity generation, it already plays the most important role in setting electricity prices
\item Poland is extreme with 72\% of fossil fuel generation and instead solar, followed by wind, being the most important sources to explain price movements
\item Neighbours are very influential in setting domestic prices. 6 out of the 11 neighbours of Germany have the German features as their dominant super-feature
\item New SSHAP explainable method is used to address the SHAP method drawbacks to deal with  high-dimensional models
\end{highlights}

\end{frontmatter}

\section{Introduction}
The European power grid is the largest grid by connected power in the world \cite{entso-e_entso-e_2025}, connecting 36 countries and managed by 40 TSOs\footnote{Each country has one TSO except Germany with five TSOs}.ENTSO-E, the European Network of Transmission System Operators for Electricity, is the association of the European transmission system operators (TSOs). The 40 TSO members are responsible for the secure and coordinated operation of Europe’s electricity system. This paper considers 27 of these countries covering 39 bidding zones with total generation of 2.7 PWh in 2025, or 8.5\% of the 32 PWh 2025 world electricity generation \cite{iea_electricity_2026}.\par
Electricity price formation in such an interconnected system is inherently complex. Prices emerge from the interaction of multiple factors, including demand, power generation mix, fuel costs, and cross-border exchanges, all of which evolve dynamically over time and space. The increasing penetration of variable renewable energy sources, such as wind and solar, further amplifies this complexity by introducing significant variability and uncertainty into the system. As a result, traditional linear or low-dimensional modelling approaches often struggle to capture the nonlinear relationships and high-dimensional dependencies that characterise modern electricity markets.\par
In recent years, machine learning techniques, and in particular deep neural networks (DNNs), have demonstrated strong performance in modelling electricity prices due to their ability to learn complex patterns from large datasets. However, their black-box nature limits their usefulness for economic interpretation and policy analysis. While these models can provide accurate predictions, they offer limited insight into the underlying drivers of price dynamics, which is a critical requirement for market participants, regulators, and system operators seeking to understand and manage increasingly complex electricity systems.\par
To address this challenge, this paper adopts an explainable artificial intelligence (XAI) approach to analyse electricity price formation across European electricity markets. By combining DNN models with SHAP-based explainability methods and reintroducing the SSHAP framework \cite{pesenti_explaining_2025}, we aim to uncover the key drivers of price dynamics while maintaining the modelling power of deep learning. In addition, we construct a synthetic EU-wide electricity market as a counterfactual scenario to explore how prices would behave under a fully integrated system with a single clearing price. This approach provides new insights into the degree of interdependence between markets and the role of interconnections in shaping electricity price formation.\par
The main contributions of this paper are as follows:
\begin{itemize}[itemsep=2pt, topsep=2pt, parsep=2pt]
    \item[$\bullet$] We analyse and compare 39 European electricity markets and create an EU electricity market to explore "What if there was a common price in the European market"
    \item[$\bullet$] We identify the most important factors driving these electricity markets from two directions, their power generation mix and their interconnected neighbours
    \item[$\bullet$] We use the SSHAP explanation method to deal with high-dimensional models
\end{itemize}
The remainder of the paper is organised as follows. Section \ref{sec:litreview} reviews the relevant literature on electricity price forecasting and the explanations of the electricity markets. Section \ref{sec:dataanalysis} describes the dataset with some exploratory analysis. Section \ref{sec:methodology} presents the models and the methodologies used in our research. Section \ref{sec:results} shows the results and Section \ref{sec:discussion} discusses them. Finally Section \ref{sec:conclusions} concludes with key findings and directions for future research.\par

\section{Literature review} \label{sec:litreview}
Electricity Price Forecasting (EPF) is a critical research area within the energy sector, driven by the inherent complexity, volatility, and deregulated nature of electricity markets \cite{maciejowska_forecasting_2022, gil_forecasting_2012}. Traditional econometric models, often referred to as white-box models, have been extensively used to predict and explain electricity prices by leveraging fundamental market variables such as supply, demand, fuel prices, and weather conditions \cite{contreras_arima_2003, conejo_forecasting_2005}.\par
A substantial body of literature documents the price-suppressing effect of renewable generation across multiple markets, including Germany \cite{wurzburg_renewable_2013}, Australia \cite{csereklyei_effect_2019}, the United States \cite{seel_impacts_2018}, Turkey \cite{acar_merit_2019}, New Zealand \cite{wen_impact_2022} and Canada \cite{stringer_power_2024}. White-box models have the advantage of providing direct explanations from their models, without intermediary methods. However while these models offer direct interpretability, they often struggle to capture the nonlinearities and high-dimensional interactions characteristic of electricity markets \cite{aggarwal_electricity_2009, weron_electricity_2014, laitsos_state_2024}.\par
For example \cite{stringer_power_2024} uses Ordinary Least Square regression models on an Ontario dataset and from the negative factors for the renewable sources derives that adding more solar and wind generation brings electricity prices down. However the R-Squared values are below 20\%, highlighting that linear models do not adequately capture the complexity of electricity systems like electricity. \cite{zakeri_role_2023} highlights the importance of natural gas in setting the electricity price in Europe by using a regression analysis based on the relationship between marginal electricity generation and prices, however the R-Squared is low again.\par
The advent of machine learning has provided new avenues to address the limitations of traditional forecasting approaches. DNNs, in particular, have shown promise in modelling complex dependencies and interactions in high-dimensional data, leading to superior predictive accuracy compared to econometric methods and simpler machine learning models \cite{khosravi_quantifying_2013, lago_forecasting_2018}. However, the black-box nature of DNNs poses significant challenges in terms of interpretability, raising concerns about trust and transparency in decision-making processes \cite{doshi-velez_towards_2017}.\par
To bridge this interpretability gap, XAI techniques have gained prominence in recent years. XAI aims to provide understandable and interpretable explanations for AI model decisions and predictions, thereby enhancing transparency, trustworthiness, and accountability \cite{bodria_benchmarking_2021, molnar_interpretable_2022, machlev_explainable_2022}. Methods such as SHAP (SHapley Additive exPlanations) and LIME (Local Interpretable Model-agnostic Explanations) have been widely adopted to quantify the contribution of individual features to specific outcomes across various sectors, including energy \cite{lundberg_unified_2017, ribeiro_why_2016}.\par
Despite the growing interest in XAI, there is a paucity of research applying these techniques to EPF models. Two notable studies \cite{heistrene_explainability-based_2023, heistrene_improved_2024} utilise density and local fit principles with explanation methods such as Permutation Feature Importance \cite{breiman_random_2001, fisher_all_2019} and SHAP to establish a trust score for each forecast, thereby improving prediction accuracy. However, these studies primarily use SHAP as intermediary variables in stacking models rather than for providing explanations of electricity market dynamics, which is the focus of our research.\par
Another relevant study \cite{trebbien_understanding_2023} employs a gradient-boosted tree model on the German electricity market using twelve variables, including five power system features, their ramps, and fuel prices. The limited number of variables allows the authors to use SHAP interaction values \cite{lundberg_consistent_2019} to analyse interactions among important variables.Three studies \cite{tschora_electricity_2022, mascarenhas_bridging_2024, pesenti_explaining_2025} utilise extensions of the open-source datasets and models developed in \cite{lago_forecasting_2021} and the SHAP method to explain the electricity market behaviours.\par
This paper extends the existing literature by focusing on providing deeper insights into multiple connected electricity markets (the European electricity markets), with explanation features based on the power generation mix and interconnected neighbours rather than lagged prices for example.\par

\section{Exploratory data analysis} \label{sec:dataanalysis}
\subsection{ENTSO-E}
Almost all the data but gas prices comes from the ENTSO-E Transparency Platform\footnote{The Dutch solar power generation data from ENTSO-E, being underreported, was replaced using data from the Dutch National Energy Dashboard at https://ned.nl/}. The gas price is the closing European benchmark TTF gas price, downloaded from yahoo finance Python API. ENTSO-E plays a crucial role in the coordination, reliability, and development of Europe's electricity transmission infrastructure. Its Transparency Platform is a publicly accessible database that provides real-time and historical data on electricity generation, load, prices, and cross-border flows. ENTSO-E represents 36 European countries, of which we selected 27 for having enough data for our research, which represent 39 bidding zones. Four countries have multiple bidding zones : Italy (6), Norway (5), Sweden (4) and Denmark (2) and two countries, Germany and Luxembourg, joined to form one bidding zone.\par
The list of selected countries with the codes of their bidding zones in brackets are: Austria (AT), Belgium (BE), Bulgaria (BG), Switzerland (CH), Czechia (CZ), Germany and Luxembourg (DE\_LU), Denmark (DK\_1, DK\_2), Estonia (EE), Spain (ES), Finland (FI), France (FR), Greece (GR), Croatia (HR), Hungary (HU), Italy (IT\_CNOR, IT\_CSUD, IT\_NORD, IT\_SARD, IT\_SICI, IT\_SUD), Lithuania (LT), Latvia (LV), the Netherlands (NL), Norway (NO\_1, NO\_2, NO\_3, NO\_4, NO\_5), Poland (PL), Portugal (PT), Romania (RO), Serbia (RS), Sweden (SE\_1, SE\_2, SE\_3, SE\_4), Slovenia (SI) and Slovakia (SK).\par
The list includes all SDAC member countries but Ireland \cite{entso-e_entso-e_2025}, and also Switzerland and Serbia. Single Day-Ahead Coupling (SDAC) is the market integration initiative in Europe that links the day-ahead electricity markets of multiple countries. It enables cross-border trading of electricity, ensuring that electricity is allocated efficiently across Europe based on supply, demand, and transmission capacity. It uses the common algorithm EUPHEMIA to calculate electricity prices and cross-border flows, maximising economic efficiency and grid stability \cite{nemos_euphemia_2024}. See the European map with bidding zones in Figure \ref{fig:dominant_gen_mix_Europe_map}.\par

\subsection{Dataset} \label{sec:dataset}
The dataset covers three years, from 1 January 2023 to 31 December 2025. While a longer time-span would have been preferable, earlier data were excluded due to the significant irregularities observed during the COVID-19 period and the subsequent energy-crisis triggered by the Russian aggression in Ukraine.\par
The variables are the actual day-ahead electricity price, load and power generation mix for each of the 39 bidding zones and gas price. The day-ahead electricity price is the most traded electricity contract according to the main European electricity market places EPEX Spot and Nord Pool. Contracts were set for one hour until September 2025 and then moved to one quarter of an hour; prices are in €/MWh. Data for load and power generation mix are covering either one hour or a quarter of an hour period, the units are MW. We only have the day-close market price for gas.\par
In this paper, the 20 power generation categories are often represented with the following 7 larger categories:
\begin{itemize}[itemsep=2pt, topsep=2pt, parsep=2pt]
    \item[$\bullet$] Solar
    \item[$\bullet$] Wind, including onshore and offshore
    \item[$\bullet$] Hydro, including run-of-river, water reservoir and pumped storage
    \item[$\bullet$] Nuclear
    \item[$\bullet$] Other renewables, including geothermal, biomass, waste and others
    \item[$\bullet$] Fossil gas
    \item[$\bullet$] Other fossil fuels, including hard and brown coal, lignite, oil and others
\end{itemize}
\begin{table}[htbp]
    \centering
    \includegraphics[width=1.0\linewidth] {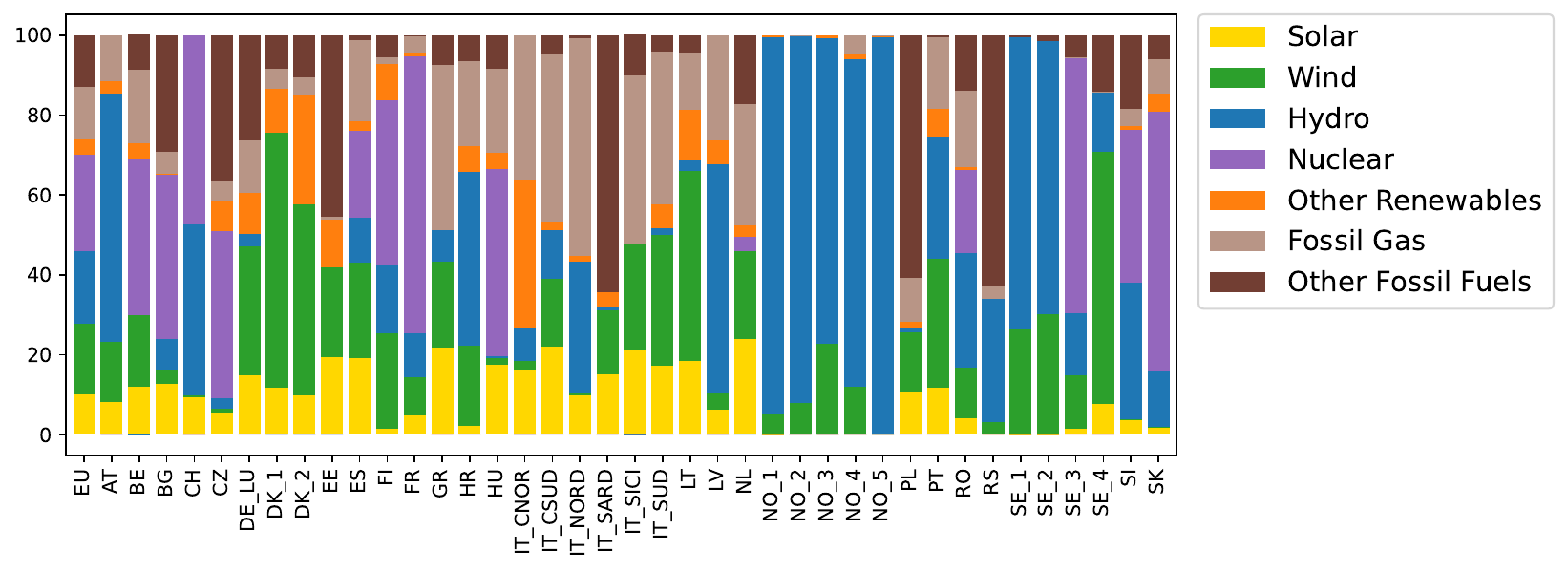}
    \caption{Power generation mix table showing high diversity between the bidding zones and solar still generating the least from the main 6 sources}
    \label{tab:gen_mix_mean_table}
\end{table}
\par
\begin{figure}[htbp]
    \centering
    \includegraphics[width=0.75\linewidth] {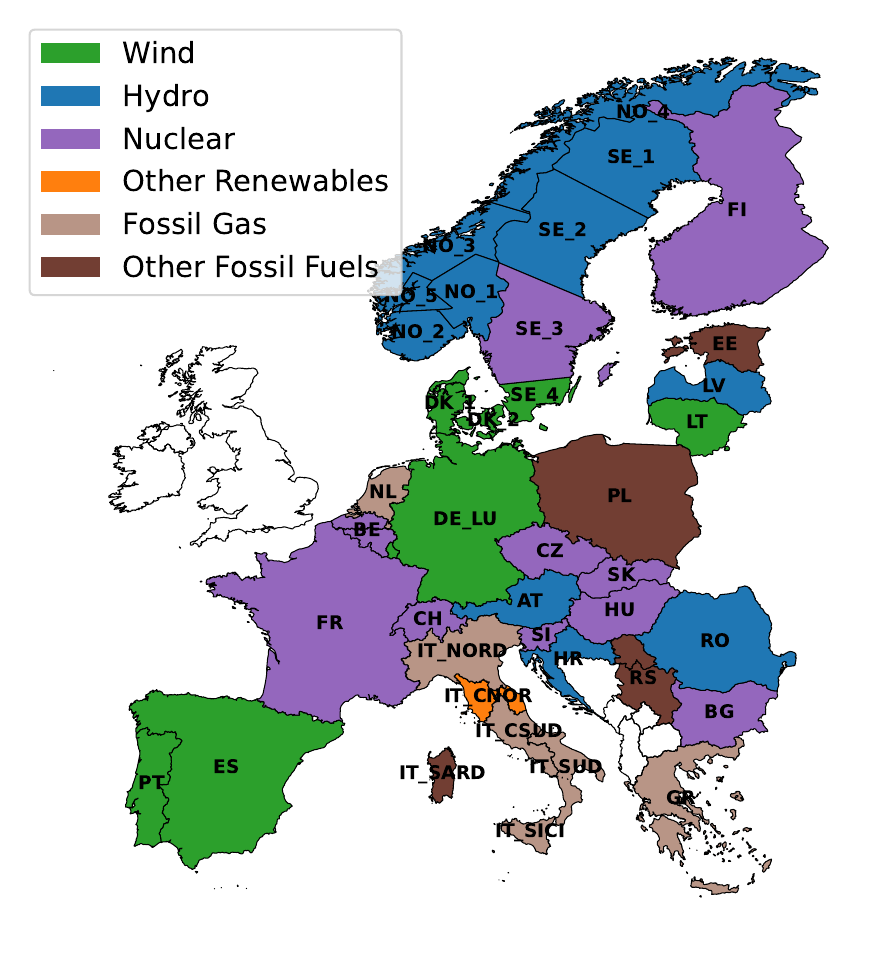}
    \caption{Dominant generation source map showing lack of geographical clusters except fuel gas for Italy and hydro for top northern bidding zones and solar being nowhere the dominant source}
    \label{fig:dominant_gen_mix_Europe_map}
\end{figure}

Table \ref{tab:gen_mix_mean_table} and the map in Figure \ref{fig:dominant_gen_mix_Europe_map} show that there is a very large diversity of power generation mix across the 39 bidding zones.\par
Overall the 39 bidding zones and the three year period, the main power generation source is from fossil fuel with 26\% of the total, of which slightly more than half is from gas, ahead of nuclear at 24\%, then wind and hydropower at both 18\%, solar at 10\% and other renewables at 4\%.\par
All the data was converted into hourly data by taking the mean of quarter of an hour data and keeping gas price constant through the day. The three years of data covers 26,304 hours, of which a total of 25,991 hours and instances were selected after cleaning the data.\par
An additional synthetic zone (EU) was constructed covering the 39 bidding zones with its loads and power generation mixes being the sums across the 39 bidding zones. Its prices are set as the weighted average of the 39 bidding zones' prices with the weights being their mean loads.\par

\subsection{Electricity prices}
\begin{figure}[htbp]
    \centering
    \includegraphics[width=1.0\linewidth] {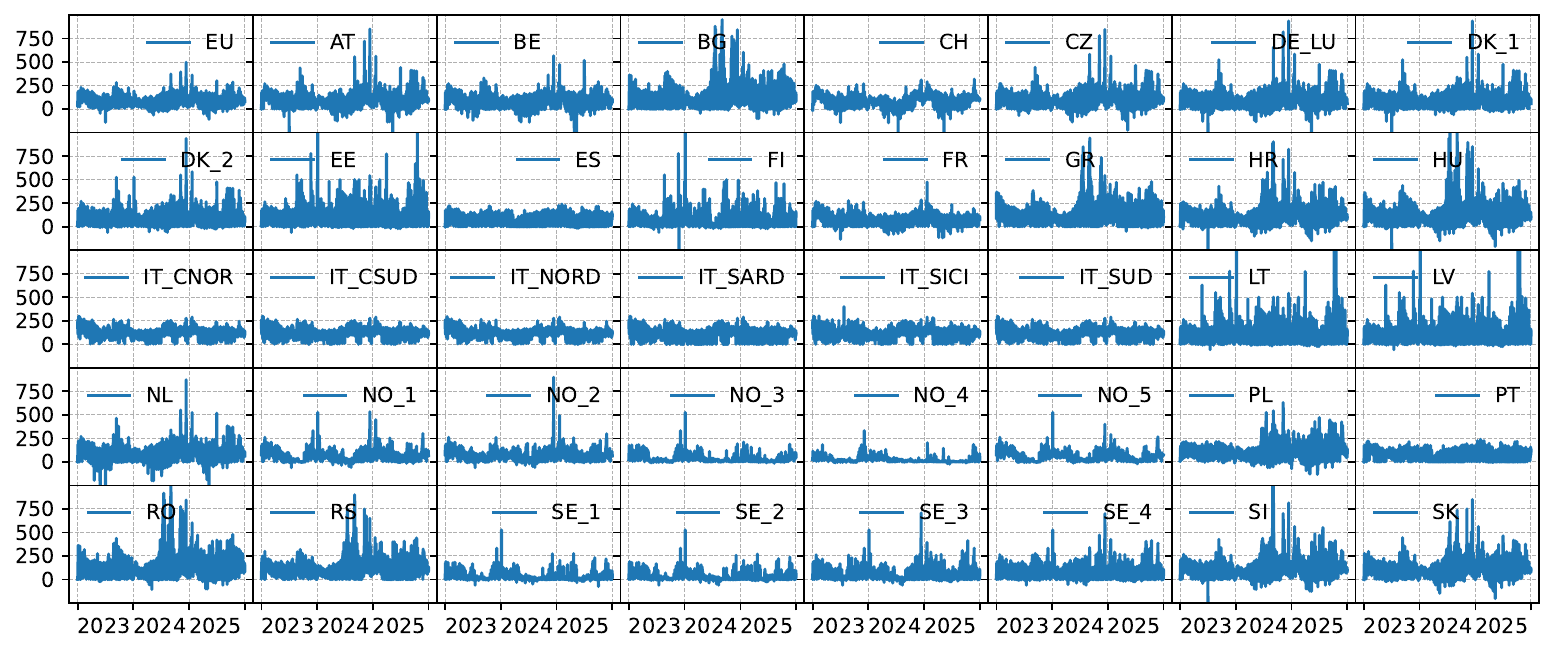}
    \caption{2023-2025 historical prices in €/MWh showing high diversity between the 40 bidding zones}
    \label{fig:40_bz_Price_plots}
\end{figure}

The integration of the European electricity markets is still ongoing, hence the 39 bidding zones show quite some diverse patterns in their historic hourly prices as we can see in Figure \ref{fig:40_bz_Price_plots}.\par
\begin{figure}[htbp]
    \centering
    \includegraphics[width=1.0\linewidth] {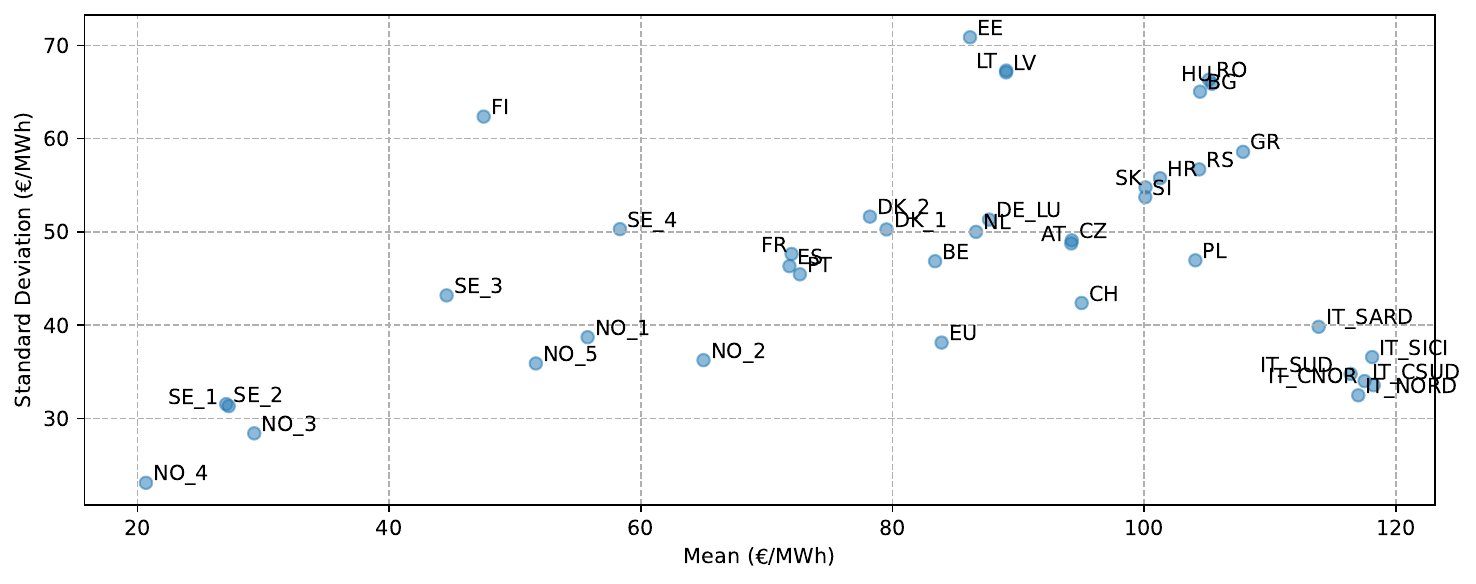}
    \caption{Mean versus standard deviation plot showing high diversity between the 40 bidding zones}
    \label{fig:mean_vs_SD_plot}
\end{figure}

Representing the 39 bidding zones with their price means and standard deviations (SD) scattered as in Figure \ref{fig:mean_vs_SD_plot}, we can identify a few clusters with different patterns:
\begin{itemize}[itemsep=2pt, topsep=2pt, parsep=2pt]
    \item[$\bullet$] Low mean and low SD for Norway and Sweden thanks to their very high level of hydropower, which has a low operational cost. The isolated further North bidding zones of Norway and Sweden have very low means and SDs while the ones further South are higher because of their interconnectors with other countries (e.g. Germany).
    \item[$\bullet$] High mean and high SD for the Baltic and Eastern European countries
    \item[$\bullet$] High mean and low SD for the six Italian bidding zones
    \item[$\bullet$] Low mean and high SD for the outlier Finland
    \item[$\bullet$] Medium mean and medium SD for the remaining bidding zones that are situated around France and Germany
\end{itemize}

\subsection{Correlation analysis} \label{sec:correlationanalysis}
\begin{figure}[htbp]
    \centering
    \includegraphics[width=1.0\linewidth] {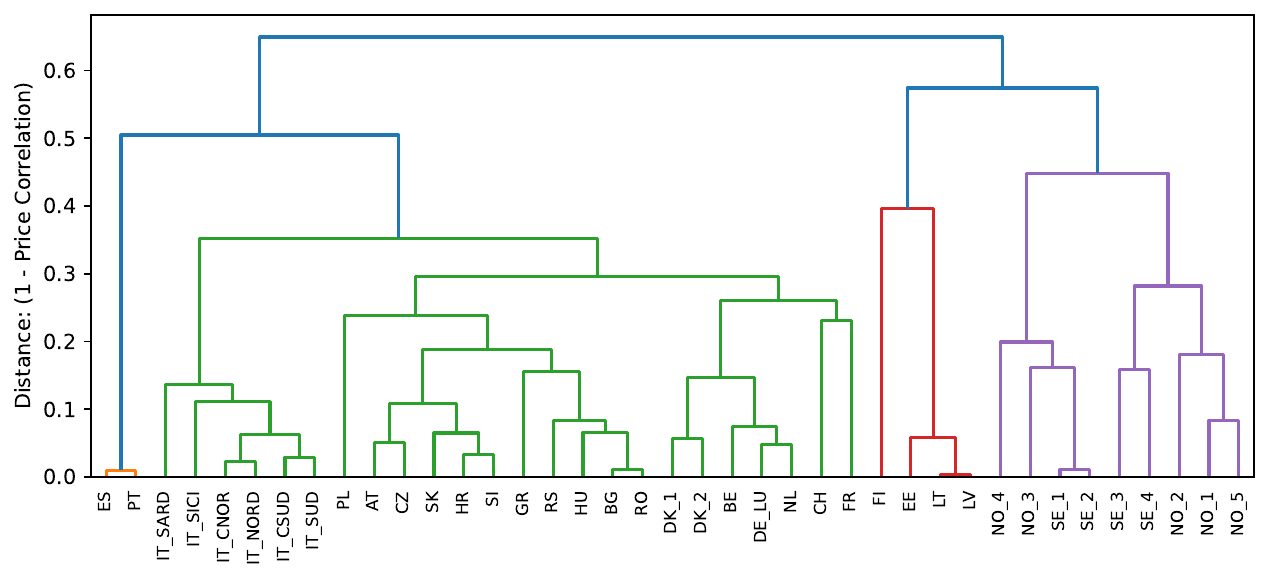}
    \caption{Hierarchical clustering dendrogram showing three small small clusters in Europe peripheries and one large main cluster around the center of Europe}
    \label{fig:clustering_dendogram}
\end{figure}

\begin{figure}[htbp]
    \centering
    \includegraphics[width=1.0\linewidth] {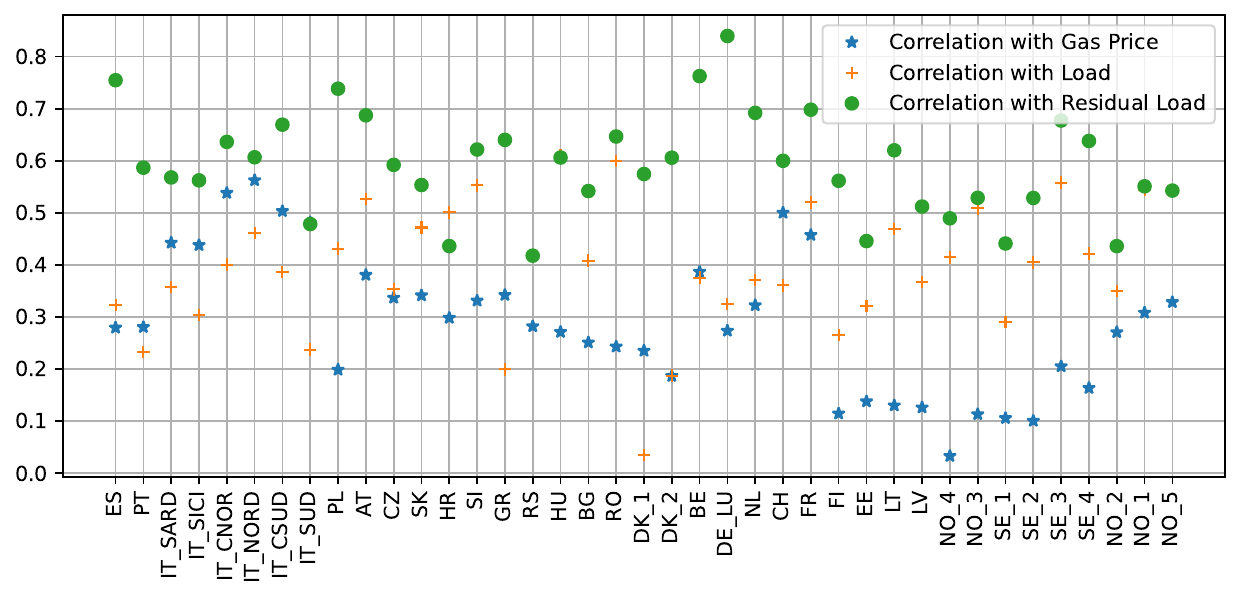}
    \caption{Correlations with electricity price plot highlighting very high values only with residual load}
    \label{fig:correlation_plot}
\end{figure}

The dendogram in Figure \ref{fig:clustering_dendogram} is based on the hierarchical clustering of the 39 bidding zones when using 1 minus price correlation as the distance metric between two bidding zones (i.e. the higher their price correlation, the closer the bidding zones). The figure highlights the following clusters with subclusters:
\begin{itemize}[itemsep=2pt, topsep=2pt, parsep=2pt]
    \item[$\bullet$] Norway and Sweden. Subclusters are the northern isolated areas and the southern areas with interconnectors to their neighbours across the seas.
    \item[$\bullet$] The Baltic countries and Finland. Lithuania and Latvia have the same prices most of the time. Finland is hardly part of the cluster.
    \item[$\bullet$] Spain and Portugal, very close between each other, and very distant from the others.
    \item[$\bullet$] The remaining of Europe with three subclusters:
    \begin{itemize}[itemsep=2pt, topsep=2pt, parsep=2pt]
        \item[$\bullet$] The six Italian bidding zones. The two islands Sicily and Sardinia are the least close.
        \item[$\bullet$] Central and Eastern Europe. Bulgaria and Romania are very close and Poland is the least close.
        \item[$\bullet$] Germany and most of its neighbours. France and Switzerland are the least close.
    \end{itemize}
\end{itemize}
Here we analyse the correlations of electricity prices with external parameters. 
As with standard competitive markets, the higher the demand, the higher the price. Hence we expect a strong correlation between electricity prices and loads. The correlation is above +20\% for all bidding zones except Denmark, see in Figure \ref{fig:correlation_plot}, and the average is +39\%. As discussed in \cite{asceric_correlation_2022}, it is better to use residual load in order to show stronger correlations. Residual load is defined as load minus solar and wind generation, i.e. load that has to be covered by "non-free" energy, as the marginal cost for solar and wind generation is almost nil. The correlation with electricity prices is as high as +84\% in Germany (as in \cite{asceric_correlation_2022}) and the average is +59\%.\par
Another strong positive correlation is expected with gas prices. This is because the electricity market uses a merit order mechanism with uniform price auction, and as analysed in \cite{gasparella_merit_2023}, gas-run power plants set the price 55\% of the time in 2022 while generating 19\% of the European electricity (and even in 2030, their projections show that the price-setting share is still close to 56\%, even with generation decreasing to 11\% of electricity). The average correlation with the 39 bidding zones is only 29\% and confirms that the relationship is more complex than linear correlation.\par

\section{Methodology} \label{sec:methodology}
\subsection{Models}
The models take load, power generation mix and gas price as input and outputs the market electricity price. The models have no time dimension, the input and output refer to the same one hour period and there is no lagged data. In fact, the day-ahead electricity price is set ahead of the actual load and power generation mix, so it is clearly not a forecasting model. The models have been built only to help to understand the mechanisms of electricity markets.\par
The main model is a straightforward multi-layer perceptron (MLP) with two hidden layers with the Adam optimiser \cite{kingma_adam_2017} and the MAE (Mean Absolute Error) loss function. The model has 128 neurons in the first hidden layer and 32 in the second one, its activation function is ReLu, the learning rate is 0.0001, the dropout rate is 0.175 and there is a L1 regularisation with lambda equal to 0.0002. The baseline models are the linear model (LR), the random forest (RF) and the gradient boosting regressor (GBR) all from the scikit-learn library\footnote{https://scikit-learn.org/stable/}.\par
The same model architecture has been used to predict each of the 39 bidding zone electricity prices. For each bidding zone model, we filter the features meaningful to the bidding zone, out of the 364 potential features: 
\begin{itemize}[itemsep=2pt, topsep=2pt, parsep=2pt]
    \item[$\bullet$] Features specific to the bidding zone: load and power generation mix
    \item[$\bullet$] Gas price
    \item[$\bullet$] Features specific to the bidding zone's neighbours. A neighbour is a bidding zone with electricity interconnectors connected to the bidding zone.
\end{itemize}    
The maximum amount of features is 122 for the Germany/Luxembourg models with 11 neighbours and the minimum is 20 for the Sicily models with one neighbour.\par
We define the EU model as the linear combination of these 39 bidding zones' models where the factors or weights are the 39 bidding zones'mean loads, which is consistent with the setting of the EU prices in Section \ref{sec:dataset}. More than being an average European price, this model was motivated by the idea of "What if there was a common price in the European market" which is a very relevant idea as Europe increases the amount of interconnection capacity.\par
The weighted average MAEs for the 39 DNN models are €12.89/MWh and €15.81/MWh for the training and test sets respectively, which compare favourably with the baseline models LR, RF and GBR with (€14.92/MWh and €16.29/MWh), (€13.00/MWh and €16.20/MWh) and (€12.43/MWh and €16.51/MWh) respectively.\par

\subsection{XAI methods} \label{sec:xaimethods}
Because these DNN models use smooth tabular data as input, all the main standard Feature Importance explainable methods gave very similar results, hence we use the SHAP method in the paper as it is the most common and it has additional properties \cite{lundberg_unified_2017}. The DNN models are trained on normalised data for inputs and outputs as standard practice. However, when applying the explainable methods, we use denormalised outputs to quantify the absolute importance of the features rather than relative values. So the SHAP values have the same unit as the electricity prices, i.e. Euros per MWh, which is useful. We use conditional expectation rather than marginal expectation when determining the Shapley values \cite{chen_true_2020, chen_algorithms_2023}. To estimate the conditional Shapley values with the DNN models, we use the Monte-Carlo sampling method developed by \cite{strumbelj_explaining_2014} and implemented in the GitHub SHAP library\footnote{https://github.com/shap/shap}. For the ensemble tree models (LR and GBR), we use Tree SHAP \cite{lundberg_consistent_2019}, which is a fast and exact algorithm to compute SHAP values.\par
SHAP values are local explanations where each value indicates the importance of a specific feature for a specific prediction. In order to get global explanations per feature, we take the mean across the dataset of the absolute values of the SHAP values. As often with many machine learning models, the overall number of features is very large, so we partition the overall set of features in a small number of subsets to become meaningful categories. The drawback of summing the mean absolute SHAP values across the subsets is to potentially overstate the importance of certain subsets where features offset each others, which is very likely when there are so many features and many have dependencies with each others. Hence we will use Super-SHAP (SSHAP) values for feature partitions or super-features as introduced in \cite{pesenti_explaining_2025} that allow the nettings to be done within the subsets before considering the global explanations (with the mean of the absolute values). A notable property of Shapley and SHAP values is the efficiency property \cite{roth_value_1988} and this is maintained in SSHAP values.\par
With SSHAPs, we will analyse the electricity markets through two different angles: the power generation mix and the interconnected neighbours. In the first case, we partition the set of 364 features in 9 super-features: the 7 power generation mix categories as in Section \ref{sec:dataset}, load and gas price. For example, the super-feature wind contains all the features onshore and offshore wind for the 39 bidding zones. We will use the same partition for the SHAP summary table \ref{tab:DNN2_SHAP_summary_table}. In the second case, we partition the set of 364 features in 40 super-features, one for each of the 39 bidding zone, and gas price. For example, the super-feature France (FR) contains the French load feature and all the French power generation mix features.\par
Finally as the EU model is a linear combination of 39 models and SHAP and SSHAP are linear functions, SHAP and SSHAP values of the EU model are simply the weighted average of the 39 model SHAP and SSHAP values. And the EU model SHAP and SSHAP values give a valuable summary of the 39 models.\par

\section{Results} \label{sec:results}
\subsection{SHAP analysis}
\begin{figure}[htbp]
    \centering
    \includegraphics[width=1.0\linewidth] {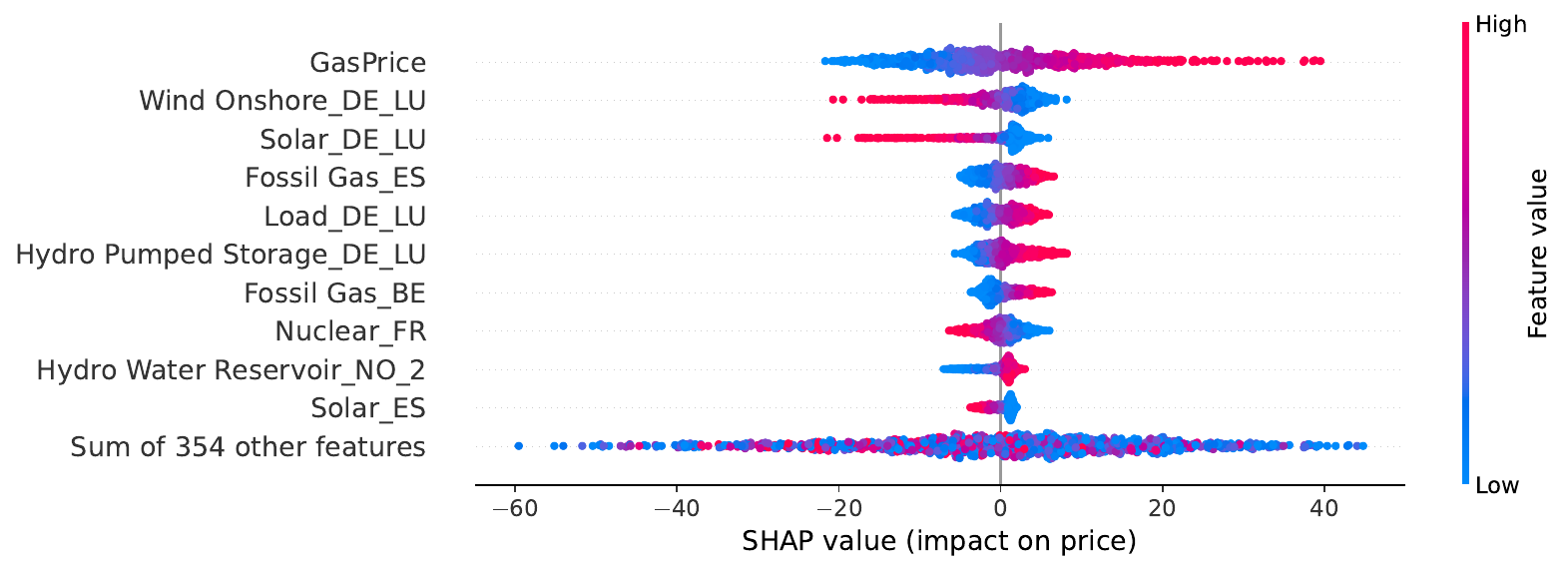}
    \caption{EU SHAP beeswarm plot showing that gas price is the most important feature by far, followed by German wind onshore and solar generation features with long negative tails}
    \label{fig:DNN2_EU_beeswarm_plot}
\end{figure}

Figure \ref{fig:DNN2_EU_beeswarm_plot} shows the SHAP beeswarm plot for the EU model with the distribution of the 10 most important features of the model, in terms of highest mean absolute SHAP values\footnote{\url{https://shap.readthedocs.io/en/latest/example_notebooks/api_examples/plots/beeswarm.html}}. Colour is used to display the original value of a feature, red being high and blue being low.\par
Table \ref{tab:DNN2_SHAP_summary_table} is the summary table of the SHAP analysis on the 40 models with the nine categories described in the previous section. "Top 80\%" in the column names means that we consider only the features with the highest mean absolute SHAP values representing 80\% of the sum of the mean absolute SHAP values. This is to ignore the low importance features which are more likely to be noisy. The last two columns estimate the importance weight of each of the nine categories and for this purpose we just consider the EU model as it contains all the features weighted by the size of the 39 bidding zones.\par
\begin{table}[htbp]
    \centering
    \includegraphics[width=1.0\linewidth] {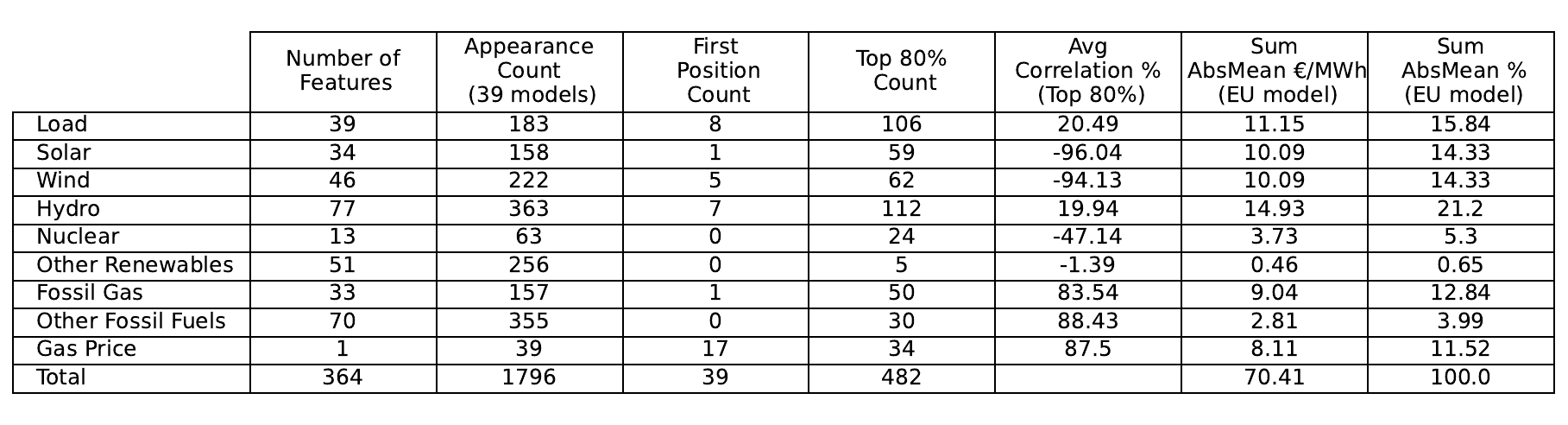}
    \caption{SHAP summary table showing hydro and load as the most two important feature categories, even if not very directional (i.e. low absolute average correlation)}
    \label{tab:DNN2_SHAP_summary_table}
\end{table}
\subsubsection{Load} \label{sec:load_SHAP}
We can see in the summary table \ref{tab:DNN2_SHAP_summary_table} that load features are used 183 times in the 39 models, 106 times they are part of the "top 80\%"  features of a model (so 77 times they are not very important) and in 8 models one load feature is the most important. Summing the mean absolute SHAP values for the 39 load features in the EU model gives €11.15/MWh, which is 15.84\% of the total and makes it the second highest across the nine categories.\par
\subsubsection{Solar and wind}
There are 34 different solar features and 46 different wind features, used 158 and 222 times respectively overall in the 39 models. Solar is once the most important feature (i.e. Spanish solar in the Portuguese model) and wind 5 times. Solar is used 59 times as "top 80\%" features, and wind 62 times. Both have very strong average correlations around -95\%, which confirms that these features are very directional. The sums of the mean absolute SHAP values for the 34 solar features in the EU model represents 14.33\% of the total and it is the same for wind.\par
\subsubsection{Hydro, nuclear and Other Renewables}
There are 77 different hydro features, 13 different nuclear features and 51 different Other Renewables features, used 363, 63 and 256 times respectively overall in the 39 models. Hydro is 7 times the most important feature and it is never the case for nuclear and Other Renewables. Hydro is used 112 times as "top 80\%" features, nuclear 24 times and Other Renewables 5 times only. The sums of the mean absolute SHAP values of the category features in the EU model represents 21.20\% of the total for hydro, the highest of all categories, 5.30\%  for nuclear and 0.65\% only for Other Renewables. For this reason, we will often skip the Other Renewables category in the remaining part of the paper.\par
\subsubsection{Fossil fuels and gas price}
There are 33 different fossil gas features, 70 different other fossil fuels features and 1 gas price feature, used 157, 355 and 39 times respectively overall in the 39 models. Fossil gas is once the most important feature and gas price is as many as 17 times the most important feature. Fossil gas is used 50 times as "top 80\%" features, other fossil fuels 30 times and gas price 34 times (i.e. only in 5 models gas price is not important). The sums of the mean absolute SHAP values of the category features in the EU model represent 12.84\% of the total for fossil gas, 3.99\%  only for other fossil fuels and 11.52\% for gas price.\par
\subsubsection{EU Model} \label{sec:EU_SHAP}
The EU model has 364 features and gas price is the most important feature by far: its mean absolute SHAP value is €8.11/MWh, which is equivalent to the sum of the second, third and fourth most important features'. Gas price does play a very important role in setting the prices in the European electricity markets as discussed in \cite{zakeri_role_2023} and Section \ref{sec:correlationanalysis}, and its top position was expected. Figure \ref{fig:DNN2_EU_beeswarm_plot} shows that it can push up the EU electricity price by as much as €40/MWh when its price is very high.\par
Wind onshore and solar powers in Germany are the second and third most important features of the EU model with their mean absolute SHAP values of €3.25/MWh and €2.69/MWh respectively. They both have asymmetrical distribution with a long tail for negative SHAP values that can push down the EU electricity price by as much as €20/MWh each when their power generation is very high. Germany has two other features in the top ten for the EU model: load and hydro pumped storage.\par
Fossil gas in Spain and Belgium are 4\textsuperscript{th} and 7\textsuperscript{th} respectively. Nuclear power in France is 8\textsuperscript{th}. Hydro water reservoir in Norway(2) is 9\textsuperscript{th}, which is notable as its capacity is only 9GW. The direction for hydro pumped storage and hydro water reservoir is opposite to wind and solar's as water is released down when electricity price is high, and pumped up when electricity price is low.\par

\subsection{SSHAP analysis with power generation mix and load}
\begin{figure}[htbp]
    \centering
    \includegraphics[width=1.0\linewidth] {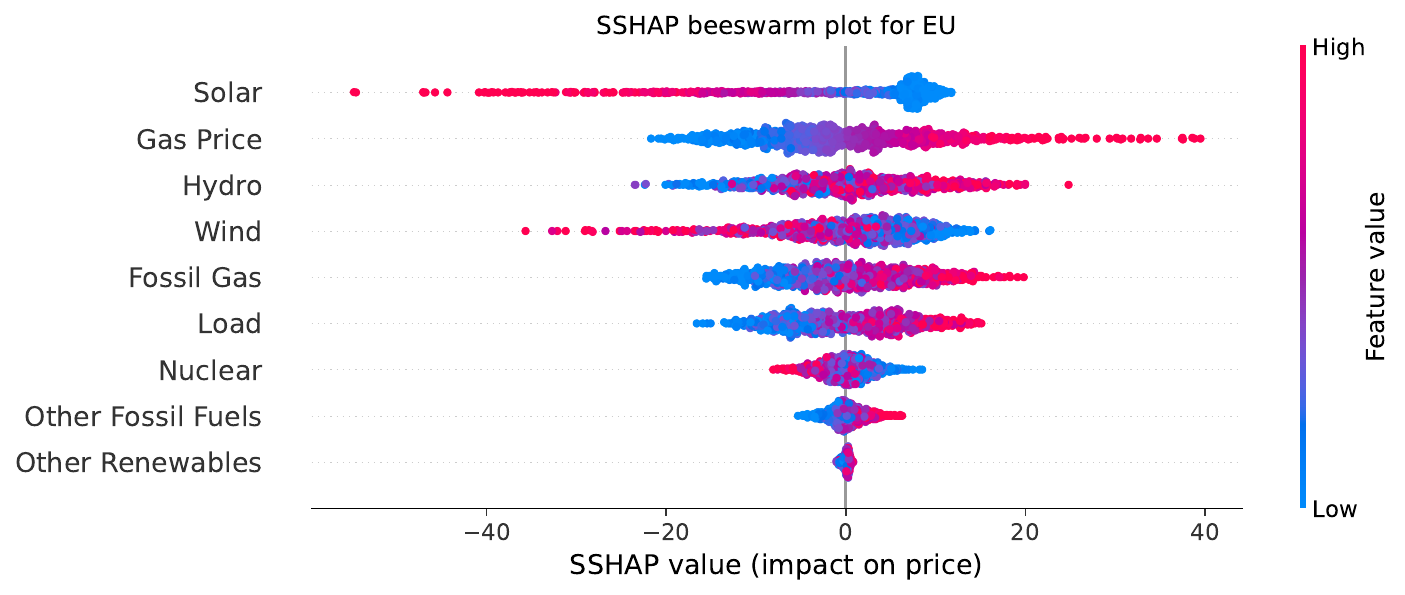}
    \caption{EU SSHAP beeswarm plot showing that solar is the most important super-feature, even if it is the least generating source, and load is now the 6\textsuperscript{th} super-feature, as many of its SHAP values have offset each others}
    \label{fig:DNN2_consolidate_EU_beeswarm_plot}
\end{figure}

Figure \ref{fig:DNN2_consolidate_EU_beeswarm_plot} shows the SSHAP beeswarm plot for the EU model with the distribution of the 9 super-features (as defined in Section \ref{sec:xaimethods}), in order of the highest mean absolute SSHAP values first. Table \ref{tab:DNN2_gen_SSHAP_relative_table} shows for each of the 40 models the relative importance of each of the 9 super-features in terms of mean absolute SSHAP values.\par
\begin{table}[htbp]
    \centering
    \includegraphics[width=1.0\linewidth] {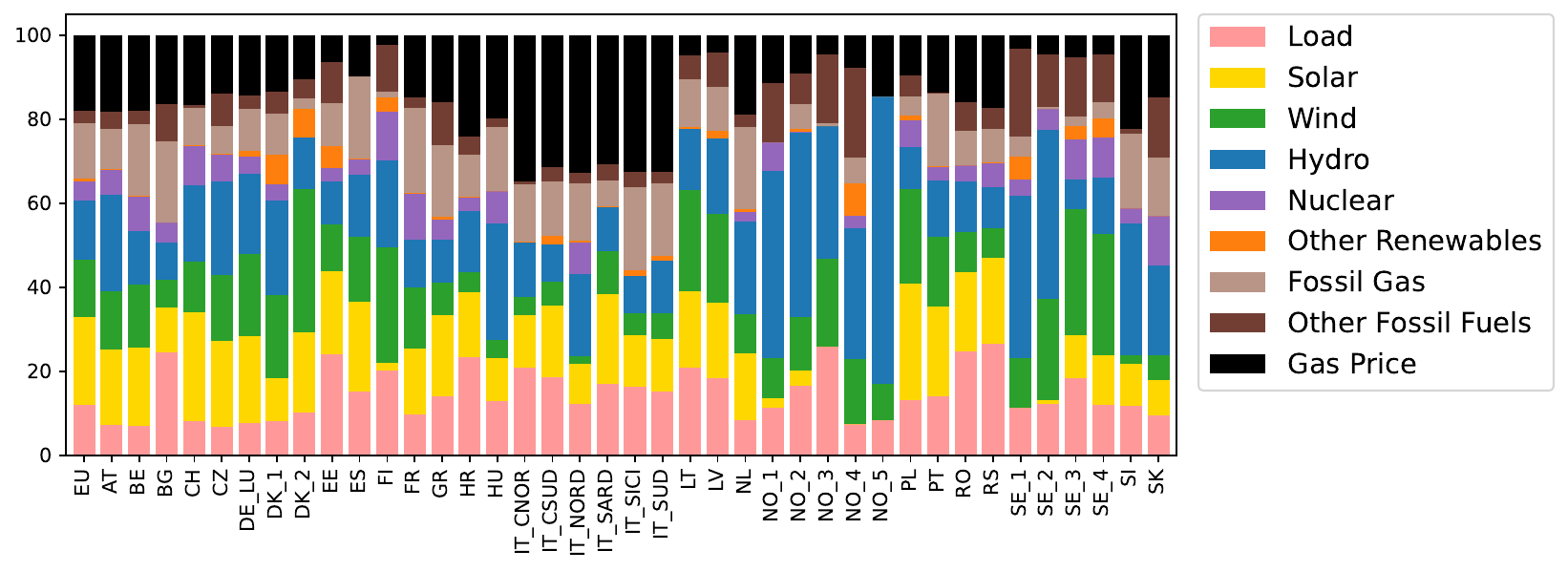}
    \caption{SSHAP summary table showing for each bidding zone the mean absolute SSHAP values for the 9 super-features}
    \label{tab:DNN2_gen_SSHAP_relative_table}
\end{table}
Solar is the most important super-feature for the EU model and its mean absolute SSHAP value is €9.42/MWh, ahead of gas price at €8.11/MWh (same as in the previous section as gas price is only one feature). Solar can push down the EU electricity price by as much as €60/MWh when its power generation is very high across Europe. Gas price can push up the EU electricity price by as much as €40/MWh when its price is very high.\par
The mean absolute SSHAP values are between €5.5MWh and €6.5MWh for hydro, wind, fossil gas and load, but only €2.11MWh for nuclear and €1.42MWh for non-gas fossil fuels.\par

\subsection{SSHAP analysis with interconnected neighbours}
\begin{figure}[htbp]
    \centering
    \includegraphics[width=0.75\linewidth] {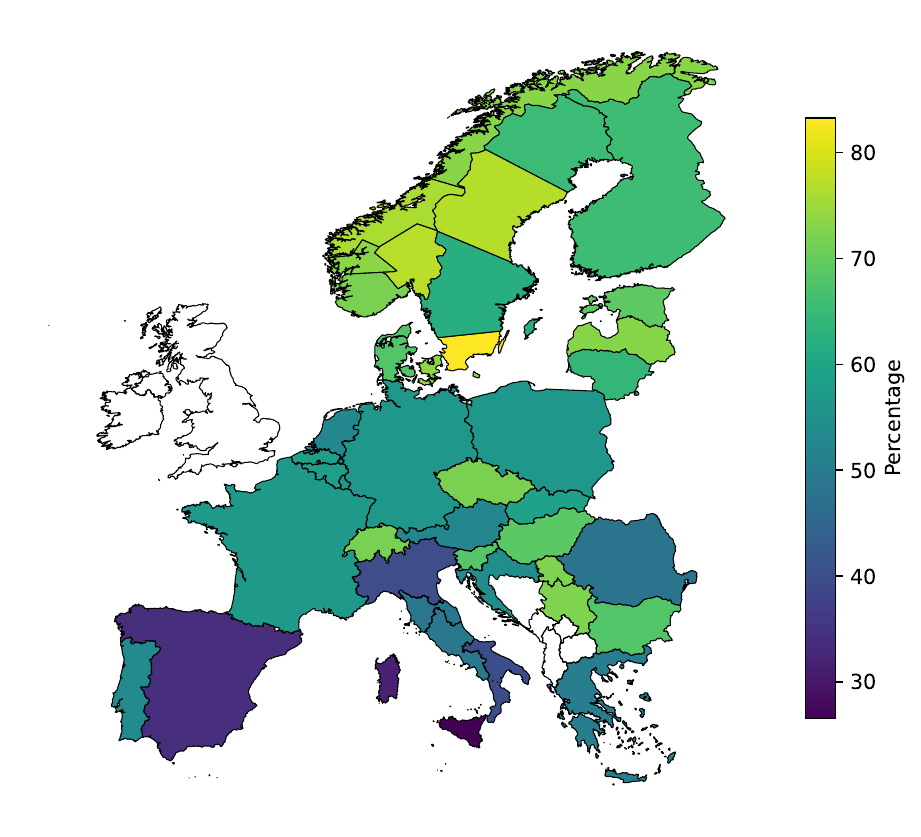}
    \caption{SSHAP neighbours importance map showing that most bidding zones have neighbours being more than 50\% responsible in the domestic price formation}
    \label{fig:neighbours_importance_Europe_map}
\end{figure}

In this section, we analyse the impact of the neighbours on the domestic electricity prices. So there are 40 super-features made of the 39 bidding zones and gas price, as described in Section \ref{sec:xaimethods}. Table \ref{tab:DNN2_bz_SSHAP_relative_table} shows for each of the model the relative importance of the super-features in terms of mean absolute SSHAP values. From Table \ref{tab:DNN2_bz_SSHAP_relative_table}, the sum of the mean absolute SSHAP values for the neighbour bidding zones are represented in the map in Figure \ref{fig:neighbours_importance_Europe_map} highlighting the importance of neighbours in setting the domestic price.\par
\begin{table}[htbp]
    \centering
    \includegraphics[width=1.0\linewidth] {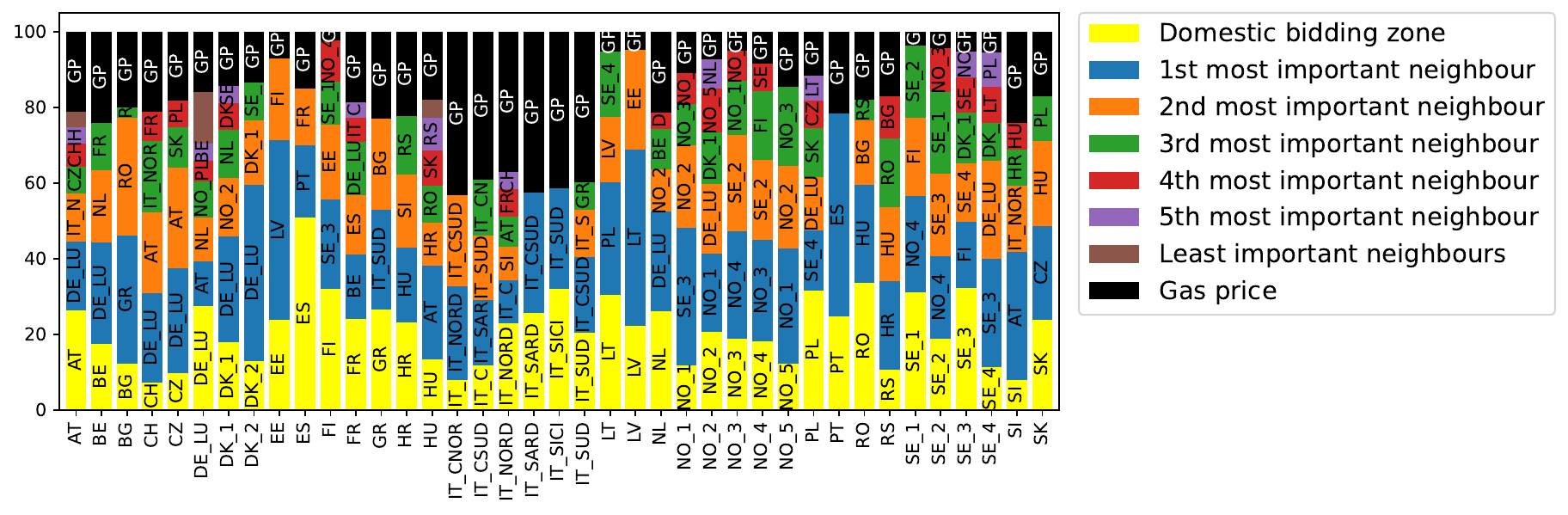}
    \caption{SSHAP summary table with bidding zone super-features where yellow shows the importance of the domestic features, black shows the importance of gas price and all the other colours show the importance of the different neighbours. The table shows that the neighbours play a very large role in setting domestic prices in most bidding zones}
    \label{tab:DNN2_bz_SSHAP_relative_table}
\end{table}
The bidding zone most dependent on neighbours is Sweden(4) with 83\% of the total mean absolute SSHAP values coming from neighbour super-features and the least dependent ones are the Italian islands and then Spain with 34\%. The average across the 39 bidding zones is as much as 61\% for the neighbour super-features. The average for the domestic super-feature is only 21\% (the remaining 18\% is for the gas price feature). Spain has the highest percentage by far with 51\% coming from its domestic super-feature, highlighting its isolation.\par

\section{Discussion} \label{sec:discussion}
\subsection{Power generation mix and load}
The SSHAP analysis reveals conclusions that differ substantially from those of the standard SHAP analysis. First, load proves less important than initially suggested. As highlighted in Section \ref{sec:load_SHAP}, the load SHAP values exhibit mixed directional impacts (both positive and negative correlations with prices), which means that some load features offset each others. By using SSHAP rather than SHAP, we get the importance after the nettings which is more revealing.\par
\begin{figure}[htbp]
    \centering
    \includegraphics[width=0.75\linewidth] {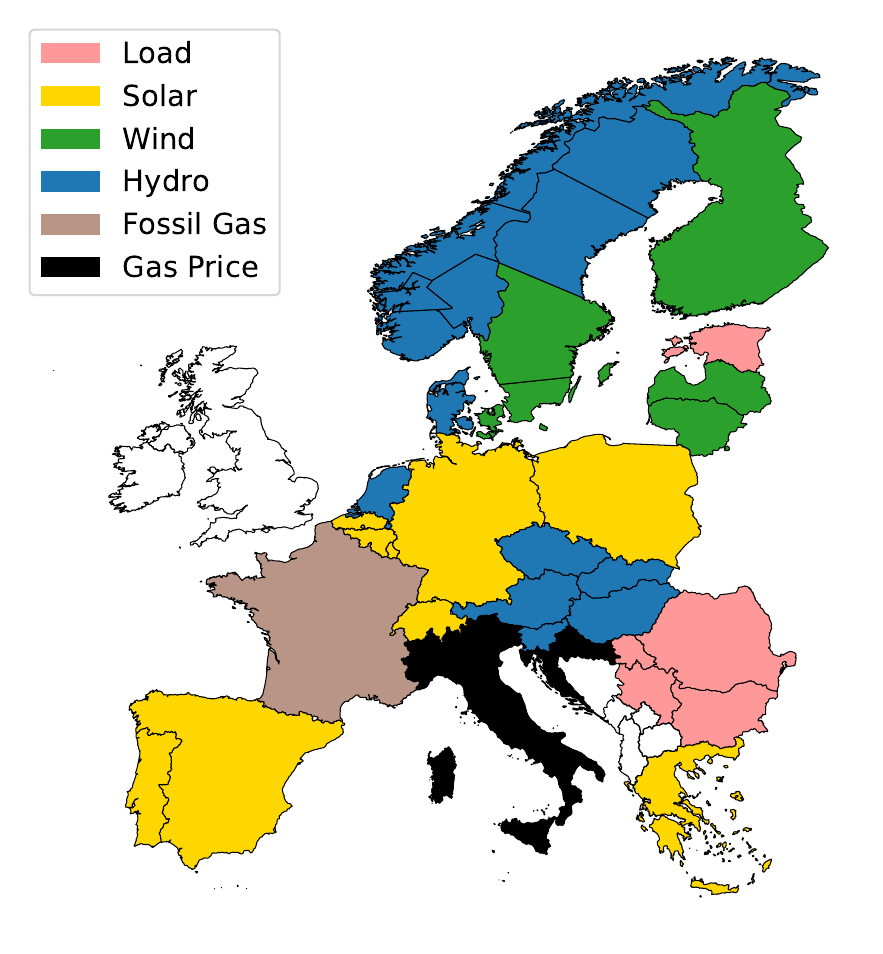}
    \caption{Dominant SSHAP super-feature map showing solar being dominant in as many as 7 medium and large bidding zones}
    \label{fig:dominant_gen_SSHAP_Europe_map}
\end{figure}
Notably, solar has the largest impact on the EU electricity prices while it is the source generating the least (excluding the Other Renewables category), see Table \ref{tab:gen_mix_mean_table}. Its mean absolute SSHAP value is €9.42/MWh, which is 30\% of the total impact from power generation sources or three times its share of EU power generation. Solar is the dominant SSHAP super-feature in 7 medium and large bidding zones (see Figure \ref{fig:dominant_gen_SSHAP_Europe_map}).\par
One reason for this is simply that the most important trait of a feature is not necessarily its mean; for example if a feature was constant, then its impact on price would be nil by definition, even if it had high values. Which is the reason why nuclear has the least impact on prices, its generation being quite stable as a base load power. Looking at standard deviation instead, solar has the highest in terms of weighted average across the 39 bidding zones of relative standard deviations at 25\%. Furthermore the total solar power generation across the 39 bidding zones, i.e. the EU solar power generation, has a standard deviation even higher relatively at 33\%. This is because solar power generations are more correlated across the 39 bidding zones than the other sources; in particular windy weather is more a localised phenomenon while sunny weather is more regional, or generally more uniform over larger areas.\par
In contrast, nuclear produces 24\% of European electricity and its EU mean absolute SSHAP value is only €2.11/MWh (7\% of total generation impact). For example France's dominant SSHAP super-feature is gas generation while it covers only 4\% of its electricity production compared to 69\% for nuclear.\par
Similarly, non-gas fossil fuels covers mainly base load. It produces 13\% of European electricity and its EU mean absolute SSHAP value is 4\% relatively. For example Poland's dominant SSHAP super-feature is solar while it covers only 11\% of its electricity production compared to 61\% for non-gas fossil fuels. Nuclear and non-gas fossil fuels are never the dominant SSHAP super-feature for any bidding zone.\par
Nordic bidding zones have their dominant SSHAP super-features, hydro and wind, matching their dominant generation sources for most of them. The exceptions are when the dominant generation source is nuclear (Sweden(3) and Finland) and the corresponding dominant SSHAP super-feature is wind and the third exception is for Denmark(1) where the dominant SSHAP super-feature is hydro while the bidding zone has no hydropower! This is explained by the cross-border influence of German and Norwegian hydro. Denmark(1) is not unique as the Netherlands has exactly the same situation and for the same reasons. This is consistent with Section \ref{sec:EU_SHAP} that reveals that hydro pumped storage in Germany and hydro water reservoir in Norway(2) are the 6\textsuperscript{th} and 9\textsuperscript{th} most important features overall.\par
There is another cluster of countries with hydro as the dominant SSHAP super-feature made of Austria, Czechia, Hungary, Slovakia and Slovenia. Czechia and Hungary have hardly any hydropower and this time the hydropower source is mainly from Austria.\par
Finally Italy is quite unique with its six bidding zones quite uniform where gas is the dominant generation source in most of the bidding zones and the dominant SSHAP super-feature is gas price for all six bidding zones. This is consistent with Italy having the highest average electricity prices in Europe with a low standard deviation (see Figure \ref{fig:mean_vs_SD_plot}) as the renewables are hardly large enough (even at peak output) to move the prices down significantly.\par

\subsection{Interconnected neighbours}
\begin{figure}[htbp]
    \centering
    \includegraphics[width=0.75\linewidth] {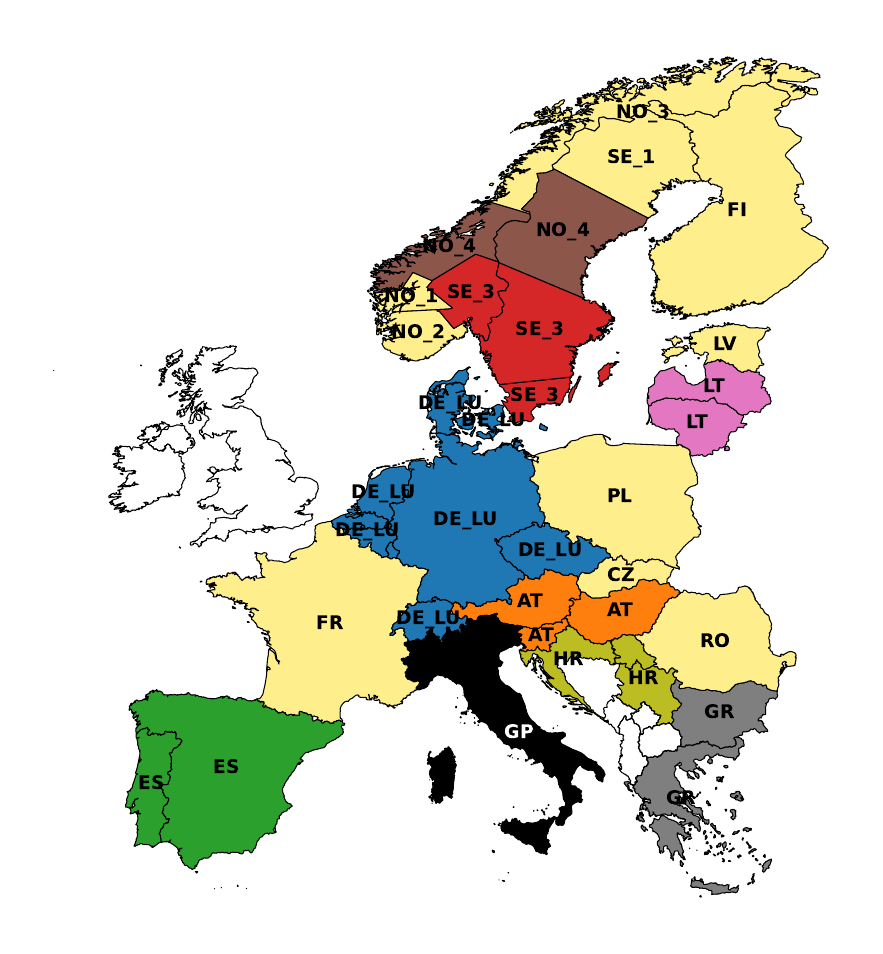}
    \caption{Dominant super-feature map highlighting that German (DE\_LU) features are the most important in 7 bidding zones. Each colour represents one specific super-feature except the cream colour representing super-features being dominant just once}
    \label{fig:dominant_bz_SSHAP_Europe_map}
\end{figure}

The map in Figure \ref{fig:dominant_bz_SSHAP_Europe_map} shows the dominant super-feature for each bidding zone. In only 13 out of the 39 bidding zones, the domestic super-feature is the most important, including the four largest bidding zones (Germany, France, Spain and Poland) as size does matter. 6 out of the 11 neighbours of Germany have the German features as their dominant super-feature, confirming the importance of Germany on its neighbours due to its large solar and wind generation capacities (52GW and 57GW respectively) and its impactful hydro pumped storage.\par
Portugal has only one neighbour, Spain, and the Spanish features are significantly more important than the Portugese features for the Portugese electricity prices (mean absolute SSHAP value for Spanish features is €24/MWh compared to €11/MWh for the Portugese features). The capacity of interconnectors between Portugal and Spain is 4.1 GW, which is 70\% of the mean load in Portugal, and helps to explain the importance of Spanish features in the Portugese electricity market. Instead 4.1 GW is 15\% only of the mean load in Spain, and the interconnector capacity with the other neighbour, France, is even lower. So the total interconnector capacity of Spain is only 29\% of its mean load and this is one reason why Spain has one of the lowest neighbour dependence percentage as set in Figure \ref{fig:neighbours_importance_Europe_map}. At the other extreme, Sweden(4) has the highest neighbour dependence percentage and its interconnector capacity is as high as 313\% of its mean load. Across the 39 bidding zones, the correlation between the neighbour dependence percentages and the interconnector capacities as a percentage of mean load is +48\%, which is significant.\par

\section{Conclusions} \label{sec:conclusions}
This paper provides an analysis of the drivers and interdependencies of European electricity prices across 39 bidding zones, using DNN models combined with XAI techniques. By leveraging SHAP and the newly introduced SSHAP framework, we move beyond predictive performance to deliver interpretable insights into the mechanisms shaping electricity price formation in highly interconnected markets.\par
Several key results emerge from the analysis. First, while load is traditionally viewed as a primary driver of electricity prices, its explanatory power is more limited than expected once the SSHAP method is used to reduce the impact of noisy features. In contrast, renewable energy sources, particularly solar, play a disproportionately large role in price formation. Despite representing a smaller share of total power generation, solar energy emerges as the most influential generation source in the SSHAP analysis, reflecting its high variability and strong directional impact on prices. Second, gas price remains a dominant and highly consistent driver across electricity markets, confirming the continued importance of gas-fired generation in the merit-order mechanism. However, its influence varies significantly depending on the power generation mix of each bidding zone, with reduced impact in systems dominated by non-gas base load sources such as hydropower. Third, the analysis highlights the critical role of interconnections. Electricity prices are not solely determined by domestic conditions: neighbouring electricity markets exert substantial influence, often exceeding that of local variables. In only a minority of bidding zones is the domestic system the primary determinant of prices. This finding underscores the high degree of interdependence within the European electricity system and the importance of cross-border flows in price formation.\par
Future work could refine the representation of interconnections by explicitly modelling transmission constraints and flows, rather than relying solely on neighbouring features. This would help to better capture congestion effects and the physical limits of market integration. Future work could also differentiate explanations for different price ranges and for very high prices in particular. As highlighted with the beeswarm plots, solar and wind distributions have long negative tails to push down prices when their generations
are very high, but they do not push much the prices up when their generations are very low or nil. Hence they are not good explanations for extreme high prices. This is confirmed by their SSHAP values being significantly lower with tree ensemble baseline models (RF and GBR)
that use RMSE (Root Mean Squared Error) as loss function rather than MAE.\par
Finally, from a policy perspective, the insights provided by this analysis could be used to assess the impact of different market design choices, such as increased interconnection capacity or alternative pricing mechanisms. Exploring counterfactual scenarios—such as a fully integrated European electricity market with a single clearing price—remains a particularly promising direction for future research.\par

\section*{Declaration of Competing Interest}
The authors declare that they have no known competing financial interests or personal relationships that could have appeared to influence the work reported in this paper.

\section*{Data Availability}
The data that support the findings of this study are available at \url{https://transparency.entsoe.eu/} and can be accessed with the code from \url{https://github.com/EnergieID/entsoe-py}

\section*{Acknowledgments}
This work was supported by EDF Energy R\&D UK and EPSRC under Grant EP/V519625/1.

\section*{Declaration of generative AI and AI-assisted technologies in the manuscript preparation process}
During the preparation of this work, the author(s) used Mistral and Claude for correcting some sections of the Python code and improving the readability of some sections of the paper. The author(s) reviewed and edited the output as needed and take full responsibility for the content of the published article.

\bibliography{references}
\end{document}